\documentclass{article}

\usepackage{microtype}
\usepackage{graphicx}
\usepackage{subfigure}
\usepackage{booktabs} 
\usepackage{amssymb}
\usepackage{scrextend}
\usepackage{amsmath}
\usepackage{multirow}
\usepackage{tabularx}
\usepackage{caption}
\usepackage{makecell}
\usepackage{enumitem}

\usepackage{hyperref}

\newcommand{\floor}[1]{\lfloor #1 \rfloor}

\usepackage[accepted]{icml2020}

\icmltitlerunning{Software Engineering Event Modeling using Relative Time in Temporal Knowledge Graphs}

\begin{document}

\twocolumn[
\icmltitle{Software Engineering Event Modeling \\ using Relative Time in Temporal Knowledge Graphs}



\icmlsetsymbol{equal}{*}

\begin{icmlauthorlist}
\icmlauthor{Kian Ahrabian}{mcgill}
\icmlauthor{Daniel Tarlow}{google,mcgill}
\icmlauthor{Hehuimin Cheng}{mcgill}
\icmlauthor{Jin L.C. Guo}{mcgill}
\end{icmlauthorlist}

\icmlaffiliation{mcgill}{School of Computer Science, McGill University, Montreal, Canada}
\icmlaffiliation{google}{Google Research, Brain Team, Montreal, Canada}

\icmlcorrespondingauthor{Kian Ahrabian}{kian.ahrabian@mail.mcgill.ca}

\icmlkeywords{Temporal Knowledge Graphs, Software Engineering}

\vskip 0.3in
]



\printAffiliationsAndNotice{}  

\begin{abstract}
We present a multi-relational temporal Knowledge Graph based on the daily interactions between artifacts in GitHub, one of the largest social coding platforms.
Such representation enables posing many user-activity and project management questions as link prediction and time queries over the knowledge graph.
In particular, we introduce two new datasets for \textit{i)} interpolated time-conditioned link prediction and \textit{ii)} extrapolated time-conditioned link/time prediction queries, each with distinguished properties.
Our experiments on these datasets highlight the potential of adapting knowledge graphs to answer broad software engineering questions. Meanwhile, it also reveals the unsatisfactory performance of existing temporal models on extrapolated queries and time prediction queries in general.
To overcome these shortcomings, we introduce an extension to current temporal models using relative temporal information with regards to past events.
\end{abstract}

\section{Introduction}
Hosting over 100 million repositories, GitHub (GH) is one of the biggest social coding platforms~\cite{githubBlog:online}.
Over the past decade, the available artifacts hosted on GH have become one of the most important resources for software engineering (SE) researchers to study various aspects of programming, software development, the characteristics of open source users and ecosystem~\cite{cosentino2017systematic}.
Example questions of interest include when an issue will be closed~\cite{kikas2016using, rees2017better}, how likely a pull request will be merged and when~\cite{gousios2014dataset, soares2015acceptance}, and who should review a pull request~\cite{yu2016reviewer, hannebauer2016automatically}.

Our aim in this work is to connect the above SE research questions to the literature on learning knowledge graph (KG) embeddings, with a particular emphasis on temporal KGs due to the importance of the temporal component in the above questions.
Methods for time prediction and time-conditioned link prediction in KGs \citep[][Section~5.1-5.2]{kazemi2020representation} are generally based on point process models \cite{trivedi2017know,trivedi2019dyrep,knyazev2019learning} or adaptations of KG embeddings that additionally use time to compute scores \cite{dasgupta2018hyte,leblay2018deriving,garcia2018learning,goel2019diachronic}.
While point processes are elegant, they are more challenging to work with and require strong assumptions on the underlying intensity functions \citep[see e.g.,][~Equation~1 \& Section~4]{trivedi2019dyrep}. 
Thus we focus on KG embedding-based methods, in particular starting with Diachronic Embeddings (DE-$X$)~\cite{goel2019diachronic} for time-varying embeddings and RotatE~\cite{sun2019rotate} for scoring facts.


Our contributions are the following:
\begin{itemize}
    \item Collecting two new temporal KG datasets\footnote{\url{https://zenodo.org/record/3928580}} from GH public events that allow casting the above SE questions as time-conditioned prediction queries (Section \ref{sec:dataset}).
    \item Benchmarking existing temporal KG embedding methods on the new datasets (Section \ref{sec:experiments}).
    \item Based on the observation that existing temporal KG embedding models do not well capture patterns in relative time that are important to SE applications, particularly in extrapolated settings, e.g., \textit{How long will it take from pull request being opened to being closed?}, we propose a new relative temporal KG embedding inspired by the use of \emph{relative time} in attention-based neural networks like Music Transformer \cite{huang2018music} and Transformer-XL \cite{dai2019transformer} (Section \ref{sec:method}).
\end{itemize}
In total, our work brings together SE research questions and temporal KG models by introducing new datasets, benchmarking existing recent methods on the datasets, and suggesting a new direction of leveraging Transformer-style relative time modeling in KG embeddings.

\begin{table*}[ht]
\caption{Characteristics comparison of the introduced datasets to existing temporal KG datasets.}
\label{tab:datasets_statistics}
\vskip 0.15in
\begin{center}
\begin{small}
\begin{sc}
\begin{tabular}{lccccccccc}
\toprule
\textbf{Dataset}                                & $\mathbf{|V|}$    & $\mathbf{|E|}$    & $\mathbf{|R|}$    & $\mathbf{|T|}$    & $\mathbf{D_{MAX}}$    & $\mathbf{D_{MED}}$    \\
\midrule
GitHub-SE 1Y                                    & 125,455,982       & 249,375,075       & 19                & 365               & 3,519,105             & 2                     \\
GitHub-SE 1Y-Repo                               &  133,335          & 285,788           & 18                & 365               & 73,345                & 1                     \\
\midrule
GitHub-SE 1M                                    & 13,690,824        & 23,124,510        & 19                & 31                & 1,324,179             & 2                     \\
GitHub-SE 1M-Node                               & 139,804           & 293,014           & 14                & 31                & 639                   & 2                     \\
\midrule
GitHub (Original)~\cite{trivedi2019dyrep}       & 12,328            & 771,214           & 3                 & 366               & -                     & -                     \\
GitHub (Subnetwork)~\cite{knyazev2019learning}  & 284               & 20,726            & 8                 & 366               & 4,790                 & 53.5                  \\
ICEWS14~\cite{garcia2018learning}               & 7,128             & 90,730            & 230               & 365               & 6,083                 & 3                     \\
ICEWS05-15~\cite{garcia2018learning}            & 10,488            & 461,329           & 251               & 4017              & 52,890                & 5                     \\
YAGO15K~\cite{garcia2018learning}               & 15,403            & 138,056           & 34                & 198               & 6,611                 & 5                     \\
Wikidata~\cite{leblay2018deriving}              & 11,153            & 150,079           & 96                & 328               & 586                   & 5                     \\
GDELT~\cite{leetaru2013gdelt}                   & 500               & 3,419,607         & 20                & 366               & 53,857                & 10,336                \\
\bottomrule
\end{tabular}
\end{sc}
\end{small}
\end{center}
\vskip -0.1in
\end{table*}

\section{Dataset}
\label{sec:dataset}

\textbf{Creation} To create the dataset, we retrieved from GH Archive\footnote{\url{https://www.gharchive.org}} all of the raw public events in GH in 2019.
The knowledge graph was then constructed by tuples, each of which represents an individual event containing temporal information based on its type and a predefined set of extraction rules. The properties of the constructed KG is shown in the first row of Table~\ref{tab:datasets_statistics}, referred to as \textsc{GitHub-SE 1Y}.

Due to the substantial size of the \textsc{GitHub-SE 1Y} and unrealistic computational demands of training KG embedding models on this KG, we sampled the \textsc{GitHub-SE 1Y} using two distinct strategies, described in more details in Appendix \ref{sup:ds:sampling}.
The first strategy aims to retain maximum temporal information about particular SE projects.
To achieve this, first, an induced sub-graph containing all related nodes was extracted for each node with type \textit{Repository}.
Then, for each sub-graph a popularity score was calculated as $P(G) = W_1 \times S_G + W_2 \times T_G$ where $S_G$ is the size of the graph, $T_G$ is the time-span of the graph, and $W_1,W_2 \in \mathbb{R}^+$ are weight values.
Finally, from the top three ranked repositories, we selected the \textit{Visual Studio Code} repository to extract a one-year slice as it exercised more functionalities related to the target entities in this work, i.e. issues and pull requests.
We name this dataset \textsc{GitHub-SE 1Y-Repo} due to its repository-centric characteristics.

The second strategy aims at preserving the most informative nodes regardless of their type. We used a variation of Snowball Sampling~\cite{goodman1961snowball} on all the events in December 2019. This sampled dataset, i.e. \textsc{GitHub-SE 1M-Node}, captures events across various projects and therefore, can be used to answer queries such as which project does a user start contributing at a certain time.

\textbf{Characteristics} In Table \ref{tab:datasets_statistics}, we compare the variations of \textsc{GitHub-SE} KG proposed in this work with commonly used datasets in the literature. Even the sampled down versions of our datasets are considerably larger in terms number of nodes. They have much higher edge to node ratios which translates into sparsity in graphs, but this sparsity level is close to what appears in GitHub as a whole.
Additionally, similar to relations, each node in our datasets is also typed.

\citet{trivedi2019dyrep} also collects a temporal KG dataset from GitHub. However, this dataset is exclusively focused on the social aspects of GitHub, discarding repositories and only including user-user interactions, and it does not appear to be publicly available beyond raw data and a small subnetwork extracted in a follow-up work \citep{knyazev2019learning}. To differentiate our datasets, which focus on the SE aspects of GitHub, we append \texttt{-SE} to the dataset names.

The distinguishing characteristics of the proposed datasets, i.e. size, sparsity, node-typing, diversity, focus on SE aspects, and temporal nature, introduce a variety of engineering and theoretical challenges that make these datasets a suitable choice for exploring and exposing the limitations of temporal knowledge graph embedding models.


\section{Method}
\label{sec:method}

\subsection{Existing KG embedding models}
We first examine the performance of the state-of-the-art KG embedding models on the \textsc{GitHub-SE 1M-Node} and \textsc{GitHub-SE 1Y-Repo}  datasets. We select RotatE~\cite{sun2019rotate} for the static settings considering its ability to infer \textit{Symmetry}, \textit{Antisymmetry}, \textit{Inversion}, and \textit{Composition} relational patterns. Moreover, we use DE-$X$~\cite{goel2019diachronic} for the dynamic setting due to its superior performance on existing benchmarks and the fact that for any static model $X$ there exists an equivalent DE-$X$ model ensuring the ability to learn aforementioned patterns.

Notationally, a KG $G$ is a set of tuples of the form
$x = (s, r, o, t)$ respectively representing the subject, relation, object, and timestamp. 
The diachronic embeddings are defined as
\begin{equation}
    D(e, t) = E(s) \oplus (E_A(e) + sin(t * E_F(e) + E_\phi(e))) \nonumber
\end{equation}
where $E, E_A, E_F, E_\phi$ are embedding lookup tables and the last three respectively represent \textit{Amplitude}, \textit{Frequency}, and \textit{Phase} of a sinusoid.
Similar to \citet{goel2019diachronic}, whenever $t$ consists of multiple set of numbers rather than one, e.g. year, month, and day, for each set of numbers a separate $D$ is defined, and the values are summed up.
Subsequently, the scoring function of the DE-RotatE model is defined as
\begin{equation}
    score(s, r, o, t) = D(s, t) \circ E_R(r) - D(o, t) \nonumber
\end{equation}
where $E_R$ is an embedding lookup table for relations.

\subsection{Relative Temporal Context}
The idea of using relative temporal information has been successfully employed in natural language processing~\cite{vaswani2017attention,dai2019transformer} and music generation~\cite{huang2018music}.
These models formalize the intuition that temporal spacing between events is more central than the absolute time at which an event happened.
We believe this framing is also appropriate for SE applications of temporal knowledge graphs: to predict if a pull request is closed at time $t$, it is more important to know how long it has been since the pull request was opened than it is to know $t$.

A challenge is that there are a lot of events, and we do not want to hard-code which durations are relevant.
Instead, we would like the model to learn which temporal durations are important for scoring a temporal fact.
As the number of related facts to an entity could be as high as a few thousand,
we propose to pick a fixed-number of facts as temporal context to provide as an input to the models.


Let $\mathcal{H}(e,r,t) = \{ t'  \mid (s', r', o', t') \in G \land (t' < t) \land \ (r' = r) \land (s' = e \lor o' = e)  \}$
be the set of times associated with facts involving entity $e$ and relation $r$ occurring before time $t$, 
and let $\delta(e, r, t) = t - \max \mathcal{H}(e,r,t)$ be the relative time since a fact involving $e$ and relation $r$ has occurred.
Hence, an entity's \emph{relative temporal context} at query time $t_q$ is $\Delta(e, t_q) = [\delta(e, 1, t_q), \ldots, \delta(e, |R|, t_q)]^\top \in \mathbb{R}^{|R|}$.


\subsection{Relative Time DE-RotatE (\textsc{RT-DE-Rotate})}
\label{subsec:rt_de_rotate}
We now turn attention to using the relative temporal context $\Delta(e, t)$ as an input to temporal KG embeddings.
Our inspiration is the \textit{Transformer} encoder, which has emerged as a successful substitute to more traditional Recurrent Neural Network approaches used for sequential ~\cite{vaswani2017attention, dai2019transformer, huang2018music} and structural~\cite{parmar2018image} tasks.
The core idea is to employ a variation of attention mechanism called \textit{Self-Attention} that assigns importance scores to the elements of the same sequence.

Unlike recurrence mechanism, the positional information is injected to the Transformer styled encoders by \textit{1)} adding sine/cosine functions of different frequencies to the input~\cite{vaswani2017attention, dai2019transformer}, or \textit{2)} directly infusing relative distance information to attention computation in form of a matrix addition~\cite{shaw2018self, huang2018music}.
\citet{vaswani2017attention} introduced a positional encoding scheme in form of sinusoidal vectors defined as $\rho(i) = [\rho_1(i), \ldots, \rho_d(i)]$, where $\rho_j(i) = \sin(i / 10000^{\frac{\floor{j/2}}{d}}) $ if $j$ is even and $\cos(i / 10000^{\frac{\floor{j/2}}{d}})$ if $j$ is odd,
$i$ is the absolute position, and $d$ is the embedding dimension. In the follow-up \textit{Transformer-XL} model, \citet{dai2019transformer} introduce a reparameterization of the relative attention where the attention score between a query element at position $i$ and a key element at position $j$ is defined as
\begin{align}
    A_{i,j}^{rel} &= \underbrace{E(i)^\top W_Q^\top W_{K,E} E(j)}_{(a)} + \underbrace{E(i)^\top W_Q^\top W_{K, \rho} \rho(i - j)}_{(b)} \nonumber \\
    &+ \underbrace{u^\top W_{K, E} E(j)}_{(c)} + \underbrace{v^\top W_{K, \rho} \rho(i - j)}_{(d)} \nonumber
\end{align}
where $E(i),E(j) \in \mathbb{R}^{d \times 1}$ are the $i$ and $j$ element embeddings, $u,v \in \mathbb{R}^{d \times 1}$, $W_Q$, $W_{K, E}$, $W_{K, \rho}$ are trainable $d \times d$ matrices, and $i - j$ is the relative position between $i$ and $j$.

\begin{table*}[!ht]
\caption{Performance comparison on time-conditioned Link Prediction. Results within the 95\% confidence interval of the best are bolded.
}
\label{tab:experiments}
\vskip 0.15in
\begin{center}
\begin{small}
\begin{sc}
\begin{tabular}{cccccccc}
\toprule
\textbf{Dataset}                    & \textbf{Type}                                 & \textbf{Model}        & $\mathbf{HITS@1}$ & $\mathbf{HITS@3}$ & $\mathbf{HITS@10}$    & $\mathbf{MR}$     & $\mathbf{MRR}$    \\
\midrule
\multirow{6}{*}{\makecell{GITHUB-SE \\1M-NODE}}  & \multirow{3}{*}{Interpolated}    & RotatE                & 47.58             & 76.66             & 88.95                 & \textbf{807.40}   & 0.6328            \\
                                    &                                               & DE-RotatE             & 47.98             & 76.92             & 88.87                 & \textbf{779.50}   & 0.6349            \\
                                    &                                               & RT-DE-RotatE (ours)   & \textbf{49.70}    & \textbf{78.67}    & \textbf{90.48}        & \textbf{773.90}   & \textbf{0.6522}   \\
\cline{2-8}
                                    & \multirow{3}{*}{Extrapolated}                 & RotatE                & 25.40             & \textbf{49.02}    & \textbf{57.54}        & \textbf{4762.87}  & 0.3797            \\
                                    &                                               & DE-RotatE             & \textbf{26.28}    & 48.53             & \textbf{57.33}        & \textbf{4840.16}  & \textbf{0.3838}   \\
                                    &                                               & RT-DE-RotatE (ours)   & \textbf{26.50}    & \textbf{49.54}    & \textbf{57.94}        & \textbf{4891.81}  & \textbf{0.3888}   \\
\midrule
\multirow{6}{*}{\makecell{GITHUB-SE \\1Y-REPO}}  & \multirow{3}{*}{Interpolated}    & RotatE                & 44.05             & 57.14             & \textbf{80.95}        & 18.54             & 0.5460            \\
                                    &                                               & DE-RotatE             & 42.17             & 53.88             & 76.88                 & 24.67             & 0.5233            \\
                                    &                                               & RT-DE-RotatE (ours)   & \textbf{48.93}    & \textbf{60.96}    & 78.32                 & \textbf{14.47}    & \textbf{0.5815}   \\
\cline{2-8}
                                    & \multirow{3}{*}{Extrapolated}                 & RotatE                & 2.11              & 4.82              & 9.71                  & 1917.03           & 0.0464            \\
                                    &                                               & DE-RotatE             & 1.77              & 4.08              & 9.10                  & 1961.75           & 0.0402            \\
                                    &                                               & RT-DE-RotatE (ours)   & \textbf{38.25}    & \textbf{40.08}    & \textbf{64.06}        & \textbf{1195.02}  & \textbf{0.4345}   \\
\bottomrule
\end{tabular}
\end{sc}
\end{small}
\end{center}
\vskip -0.1in
\end{table*}

The main difference in our setting is that the above models compute a score based on a single relative time $i - j$, while our relative temporal context $\Delta$ contains $|R|$ relative times for each entity.
Our approach is to score a tuple $(s, r, o, t)$ based on the information available at query time $t_q$.\footnote{During training, $t_q$ is the $t$ of the positive sample, and during evaluation, $t_q$ is set to the maximum timestamp in the training set.}
For each entity $e \in (s, o)$ we define a positional embeddings matrix $P$ of relative times between $t$ and the events in its relative temporal context $\Delta(e, t_q)$ as
\begin{align}
    P(e, t, t_q) = \begin{bmatrix} 
    \rho(t - t_q + \delta(e, 1, t_q)) \\ 
    \rho(t - t_q + \delta(e, 2, t_q)) \\
    \vdots \\
    \rho(t - t_q + \delta(e, |R|, t_q))
    \end{bmatrix} \in \mathbb{R}^{|R| \times d}. \nonumber
\end{align}
Intuitively, these relative times encode ``if the event happened at time $t$, how long would it have been since the events in the relative time context.''
A learned, relation-specific row vector $W_{P}(r) \in \mathbb{R}^{1\times|R|}$ for $r=1, \ldots, |R|$ chooses which rows of $P$ are important, and then $\gamma(r, e, t, t_q) = W_{P}(r) P(e, t, t_q) \in \mathbb{R}^{1 \times d}$, abbreviated $\gamma(r, e, t)$, embeds the relative temporal context of $e$, replacing $W_{K,\rho} \rho(i - j)$:
\begin{align}
    A_x^{rel} &= \underbrace{ D(s, t) W(r) D(o, t)^\top }_{(a)} + \underbrace{ \gamma(r, s, t) W_P \gamma(r, o, t)^\top }_{(d)} \nonumber \\
    &+ \underbrace{ E(s) W_E \gamma(r, o, t)^\top }_{(b)} + \underbrace{ \gamma(r, s, t) W_E^\top E(o)^\top }_{(c)} \nonumber
\end{align}
where $W(r)$ is a relation-specific weight matrix and $W_E$ and $W_P$ are tuple-agnostic weight matrices;
however, this formulation is suitable for bilinear models.
Hence, we derive a translational variation for the DE-RotatE model as
\begin{align}
    A_x^{rel} &= \underbrace{\| D(s, t) \circ E_R(r) - D(o, t) \|}_{(a)} \nonumber \\
    &+ \underbrace{\| E(s) W_E - \gamma(r, o, t) \|}_{(b)} \nonumber \\
    &+ \underbrace{\| \gamma(r, s, t) - E(o) W_E \|}_{(c)} \nonumber
\end{align}
where $W(r)$ is replaced by an embedding lookup table $E_R(r)$. Intuitively, under this formulation $(a)$ capture entities compatibility and $(b)$ and $(c)$ capture entity-specific temporal context compatibility.
In comparison, the existing models only include term $(a)$ discarding terms $(b)$ and $(c)$.

\section{Experiments}
\label{sec:experiments}

\textbf{Datasets:} We use a 90\%-5\%-5\% events split for constructing the train, validation, and test sets.
For the interpolated queries the split was done randomly, whereas we split the data using event timestamps for the extrapolated queries.
Table \ref{tab:split} in the Appendix presents details of the splits.

\textbf{Queries:} For time-conditioned link prediction, we selected events related to the resolution of Github issues and pull requests due to their direct impact on software development and maintenance practice.
Particularly, we used \textit{``Who will close issue X at time T?"} and \textit{`Who will close pull-request X at time T?"} for evaluation.
For time prediction, we used the analogous time queries of the aforementioned queries for evaluation, i.e. \textit{``When will issue X be closed by user Y?"} and \textit{`When will pull-request X be closed by user Y?"}.

\textbf{Evaluation and Results:} 
We calculated the standard metrics to evaluate the model performance on the test set. For the extrapolated time-conditioned link prediction queries, after using the validation set for hyperparameter tuning, we retrained the selected models using both training and validation sets for evaluation. We also report the model performance without retraining in the Appendix Table \ref{tab:experiments_e}.

In Table \ref{tab:experiments} we compare the model performance on the time-conditioned link prediction queries. On the \textsc{Github-SE 1M-NODE} queries, our model slightly outperforms existing models in some cases, but the difference is statistically insignificant in others. On the \textsc{Github-SE 1Y-REPO}, on the other hand, our \textsc{RT-DE-ROTATE} model shows a significant performance boost, particularly on the extrapolated time-conditioned link prediction queries, indicating the importance of using relative time as temporal context.

For the extrapolated time prediction queries on \textsc{GITHUB-SE 1Y-REPO} dataset, our model performed slightly better on HITS@1, HITS@3, and Mean Reciprocal Rank (MRR) than the other existing models while marginally surpassing the random baseline on all metrics. These results, detailed in the Appendix Table \ref{tab:experiments_t}, stress the necessity of having further studies on extrapolated time prediction queries.

\section{Conclusion}
In this work, we bridge between the SE domain questions and the literature on KG embedding models by introducing two new datasets based on the daily interactions in the GH platform and casting those questions as queries on an appropriate KG. Furthermore, we introduce RT-X, a novel extension to existing KG embedding models that make use of relative time with regards to past relevant events. Our experiments highlight shortcomings of existing temporal KG embedding models, notably on extrapolated time-conditioned link prediction, and exhibit the advantage of leveraging relative time as introduced in the RT-X model. 
In total, this work highlights new opportunities for improving temporal KG embedding models on time prediction queries.

\clearpage
\section*{Acknowledgements}
We acknowledge the support of the Natural Sciences and Engineering Research Council of Canada (NSERC). This research was enabled in part by support provided by Google, Calcul Qu\'ebec, and Compute Canada. We thank Daniel Johnson for helpful comments.

\bibliography{paper}
\bibliographystyle{icml2020}
\clearpage
\appendix

\newcolumntype{Y}{>{\centering\arraybackslash}X}

\section{Dataset}
\subsection{Sampling}
\label{sup:ds:sampling}
Algorithm \ref{alg:snowball_sampling} describes the snowball sampling used to create the \textsc{GitHub-SE 1M-NODE} dataset. This algorithm aims at preserving the most informative nodes regardless of their types.
Moreover, Algorithm \ref{alg:temporal_sampling} describes the temporal sampling used to create the \textsc{GitHub 1Y-REPO} dataset. This algorithm aims at preserving maximum temporal information regarding particular repositories.

\begin{algorithm}[h!]
\caption{Snowball Sampling strategy used for extracting the \textsc{GitHub-SE 1M-NODE} dataset.}
\label{alg:snowball_sampling}
\begin{algorithmic}
    \REQUIRE{set of nodes $V$, set of edges $E$, sample size $N$, growth size $S$, initial sample size $K$}
    \STATE{$L$ $\gets$ $sortDescending(V)$ w.r.t node degree}
    \STATE{$Q$ $\gets$ $MaxPriorityQueue()$}
    \FOR{$i = 1, ..., K$}
        \STATE{$Q$.put($L[i]$)}
    \ENDFOR
    \STATE{$U$ $\gets$ $Set()$}
    \WHILE{$size(U) < N$}
        \STATE{$V_u$ $\gets$ $Q$.top() w.r.t node degree}
        \STATE{$U$.put($V_u$)}
        \STATE{$U_s$ $\gets$ $randomSample(E[V_u])$ with size $S$}
        \FOR{$i = 1, ..., S$}
            \STATE{$Q$.put($U_s[i]$)}
        \ENDFOR
    \ENDWHILE
\end{algorithmic}
\end{algorithm}

\begin{algorithm}[h!]
\caption{Temporal Sampling strategy used for extracting the \textsc{GitHub 1Y-REPO} dataset.}
\label{alg:temporal_sampling}
\begin{algorithmic}
    \REQUIRE{set of nodes $V$, size importance factor $W_1$, time span importance factor $W_2$, sample size $N$}
    \STATE{$R$ $\gets$ $extractRelated(V)$} \COMMENT{\textit{Repository} node type only}
    \STATE{$P$ $\gets$ $Array(|R|)$}
    \FOR{$i = 1, ..., |R|$}
        \STATE{$P_i$ $\gets$ $calculatePopularity(R_i, W_1, W_2)$}
    \ENDFOR
    \STATE{$S$ $\gets$ $sorted(P)$}
    \STATE{$U \gets$ $Set()$}
    \FOR{$i = 1, ..., |S|$}
        \IF{$size(U) < N$}
            \STATE{$U$.union($S_i$)}
        \ENDIF
    \ENDFOR
\end{algorithmic}
\end{algorithm}

\subsection{Extraction}
Table \ref{tab:extraction} presents the set of extraction rules used to build the KG from raw events each representing a relation type. Although 80 extractions rules are defined in Table \ref{tab:extraction}, the raw events that we used only contained 18 of them.

The codes presented in the Relation column of Table \ref{tab:extraction}, when divided on underscore, are interpreted as \textit{a)} the first and the last components respectively represent entity types of the event's subject and object, \textit{b)} \textit{AO}, \textit{CO}, \textit{SE}, \textit{SO}, and \textit{HS} are abbreviations of extracted information from raw payloads~\footnote{\url{https://developer.github.com/webhooks/event-payloads/}} serving as distinguishers between different relation types among entities, and \textit{c)} the second to the last component represents the concrete action taken that triggers the event.

\begin{table*}
\caption{Details of train, validation, and test splits.}
\label{tab:split}
\vskip 0.15in
\begin{center}
\begin{small}
\begin{sc}
\begin{tabularx}{\linewidth}{@{}YYYYY@{}}
\toprule
\textbf{Dataset}                    & \textbf{Type}         & \textbf{\#Train}          & \textbf{\#Validation}     & \textbf{\#Test}           \\
\midrule
\multirow{4}{*}{GITHUB-SE 1M-NODE}  & Interpolated          & 285,953                   & 3,530                     & 3,531                     \\
                                    & Extrapolated          & 281,056                   & 2,104                     & 3,276                     \\
                                    & Standard Extrapolated & \multirow{2}{*}{275,805}  & \multirow{2}{*}{2,104}    & \multirow{2}{*}{3,276}    \\
\midrule
\multirow{4}{*}{GITHUB-SE 1Y-REPO}  & Interpolated          & 282,597                   & 1,595                     & 1,595                     \\
                                    & Extrapolated          & 269,789                   & 2,281                     & 1,472                     \\
                                    & Standard Extrapolated & \multirow{2}{*}{252,845}  & \multirow{2}{*}{2,281}    & \multirow{2}{*}{1,472}    \\
\bottomrule
\end{tabularx}
\end{sc}
\end{small}
\end{center}
\vskip -0.1in
\end{table*}

\section{Model}
\subsection{Complexity}
Table \ref{tab:complexity} presents time and space complexity comparison between the existing models and the introduced RT-X model. Notice that, while yielding superior performance, the number of free-parameters introduced in our extension does not increase linearly with the number of entities which is one of the bottlenecks of training large KG embedding models.

\begin{table*}[ht]
\caption{Time and space complexity comparison of the models given static embedding dimension $d_s$, diachronic embedding dimension $d_t$, relative time embedding dimension $d_r$, entity set $E$, and relation set $R$.}
\label{tab:complexity}
\vskip 0.15in
\begin{center}
\begin{small}
\begin{sc}
\begin{tabularx}{\linewidth}{@{}YYY@{}}
\toprule
\textbf{Model}      & \textbf{Computational Complexity} & \textbf{Free Parameters Complexity}           \\ \midrule
RotatE              & $O(d_s)$                          & $O(d_s(|E| + |R|))$                           \\
DE-RotatE           & $O(d_s + d_t)$                    & $O((d_s + d_t)(|E| + |R|))$                   \\
RT-DE-RotatE (ours) & $O(d_s + d_t + d_sd_r + d_r|R|)$  & $O((d_s + d_t)(|E| + |R|) + d_r^2 + d_sd_r)$  \\
\bottomrule
\end{tabularx}
\end{sc}
\end{small}
\end{center}
\vskip -0.1in
\end{table*}

\begin{table*}[ht!]
\caption{Performance comparison on standard extrapolated time-conditioned Link Prediction. Results within the 95\% confidence interval of the best are bolded.}
\label{tab:experiments_e}
\vskip 0.15in
\begin{center}
\begin{small}
\begin{sc}
\begin{tabular}{ccccccc}
\toprule
\textbf{Dataset}                    & \textbf{Model}        & $\mathbf{HITS@1}$ & $\mathbf{HITS@3}$ & $\mathbf{HITS@10}$    & $\mathbf{MR}$     & $\mathbf{MRR}$    \\
\midrule
\multirow{3}{*}{GITHUB-SE 1M-NODE}  & RotatE                & 19.60             & \textbf{38.37}    & \textbf{45.54}        & 6437.30           & 0.2965            \\
                                    & DE-RotatE             & 20.97             & \textbf{38.03}    & \textbf{45.21}        & 6504.79           & 0.3005            \\
                                    & RT-DE-RotatE (ours)   & \textbf{22.10}    & \textbf{38.61}    & \textbf{45.54}        & \textbf{5782.83}  & \textbf{0.3113}   \\
\midrule
\multirow{3}{*}{GITHUB-SE 1Y-REPO}  & RotatE                & 0.41              & 1.49              & 2.45                  & 2259.03           & 0.0141            \\
                                    & DE-RotatE             & 5.16              & 8.83              & 16.44                 & \textbf{1342.25}  & 0.0911            \\
                                    & RT-DE-RotatE (ours)   & \textbf{38.59}    & \textbf{40.01}    & \textbf{43.27}        & 1613.70           & \textbf{0.4034}   \\
\bottomrule
\end{tabular}
\end{sc}
\end{small}
\end{center}
\vskip -0.1in
\end{table*}

\subsection{Loss Function}
Similar to the self-adversarial negative sampling introduced in \citet{sun2019rotate}, we use weighted negative samples as
\begin{equation}
    p(h_j^{'}, r, t_j^{'}|\{(h_i, r, t_i)\}) = \frac{\exp \eta f_r(h_j^{'}, t_j^{'})}{\sum_i \exp \eta f_r(h_i^{'}, t_i^{'})} \nonumber
\end{equation}
where $\eta$ is the sampling temperature and $(h_i^{'}, r, t_i^{'})$ is the $i$-th negative sample. Hence, the loss function is defined as
\begin{align}
    L &= - \ \log\sigma(\omega - d_{r}(h, t)) \nonumber \\
    &- \sum_{i=1}^{n} p(h_i^{'}, r, t_i^{'})\log\sigma(d_r(h_i^{'}, t_i^{'})-\omega) \nonumber
\end{align}
where $\omega$ is a fixed margin and $\sigma$ is the sigmoid function.

\section{Experiments}

\subsection{Time Prediction}
To evaluate the time prediction queries, we consider the dates in the min-max timestamp range of the set that is being evaluated as the candidate set.

\subsection{Model Selection}
The best model is selected using the MRR on validation set and HITS@N with $N = 1, 3, 10$, Mean Rank (MR), MRR on test set are reported.

\subsection{Negative Sampling}
We follow the negative sampling scheme employed by \citet{dasgupta2018hyte} providing the model with both sets of time-agnostic and time-dependent negative samples.

\subsection{Regularization}
We apply L3 regularization parameterized by $\lambda$ as introduced in \citet{lacroix2018canonical} on $E$, $E_A$, $W_E$, and $W_P$.

\subsection{Re-ranking Heuristics}
We employed two re-ranking heuristics during the evaluation time for time-conditioned link prediction.
First, each entity was only evaluated among entities with the same type.
Next, we push down the ranks of entities with prior interactions with the given entity.

\begin{table}[ht!]
\caption{Hyperparameter ranges used for experiments.}
\label{tab:hyperparameters_range}
\vskip 0.15in
\begin{center}
\begin{small}
\begin{sc}
\begin{tabular}{cc}
\toprule
\textbf{Hyperparameter} & \textbf{Range} \\
\midrule
Dropout & $\{ 0.0, 0.2, 0.4 \}$ \\
$\eta$ & $\{ 0.5, 1.0 \}$ \\
$\omega$ & $\{ 3.0, 6.0, 9.0 \}$ \\
$\alpha$ & $\{ 10^{-3}, 10^{-4}, 3 \times 10^{-5}, 10^{-5} \}$ \\
$\lambda$ & $\{ 10^{-3}, 5 \times 10^{-4}, 10^{-4} \}$ \\
$d_s$ & $\{ 128, 96, 64, 32, 0 \}$ \\
$d_a$ & $\{ 128, 96, 64, 32, 0 \}$ \\
$d_r$ & $\{ 128, 64, 32, 0 \}$ \\
\bottomrule
\end{tabular}
\end{sc}
\end{small}
\end{center}
\vskip -0.15in
\end{table}

\begin{table}[ht!]
\caption{Average runtime comparison of the models in seconds.}
\label{tab:runtime}
\vskip 0.15in
\begin{center}
\begin{small}
\begin{sc}
\begin{tabular}{cccc}
\toprule
\textbf{Model}          & \textbf{Samples}  & \textbf{Avg Runtime}  \\
\midrule
RotatE                  & 6400              & 77s                   \\
DE-RotatE               & 6400              & 80s                   \\
\textbf{RT-DE-RotatE}   & 6400              & 87s                   \\
\bottomrule
\end{tabular}
\end{sc}
\end{small}
\end{center}
\vskip -0.15in
\end{table}

\begin{figure*}[ht!]
\vskip 0.2in
\begin{center}
\centerline{\includegraphics[width=\linewidth]{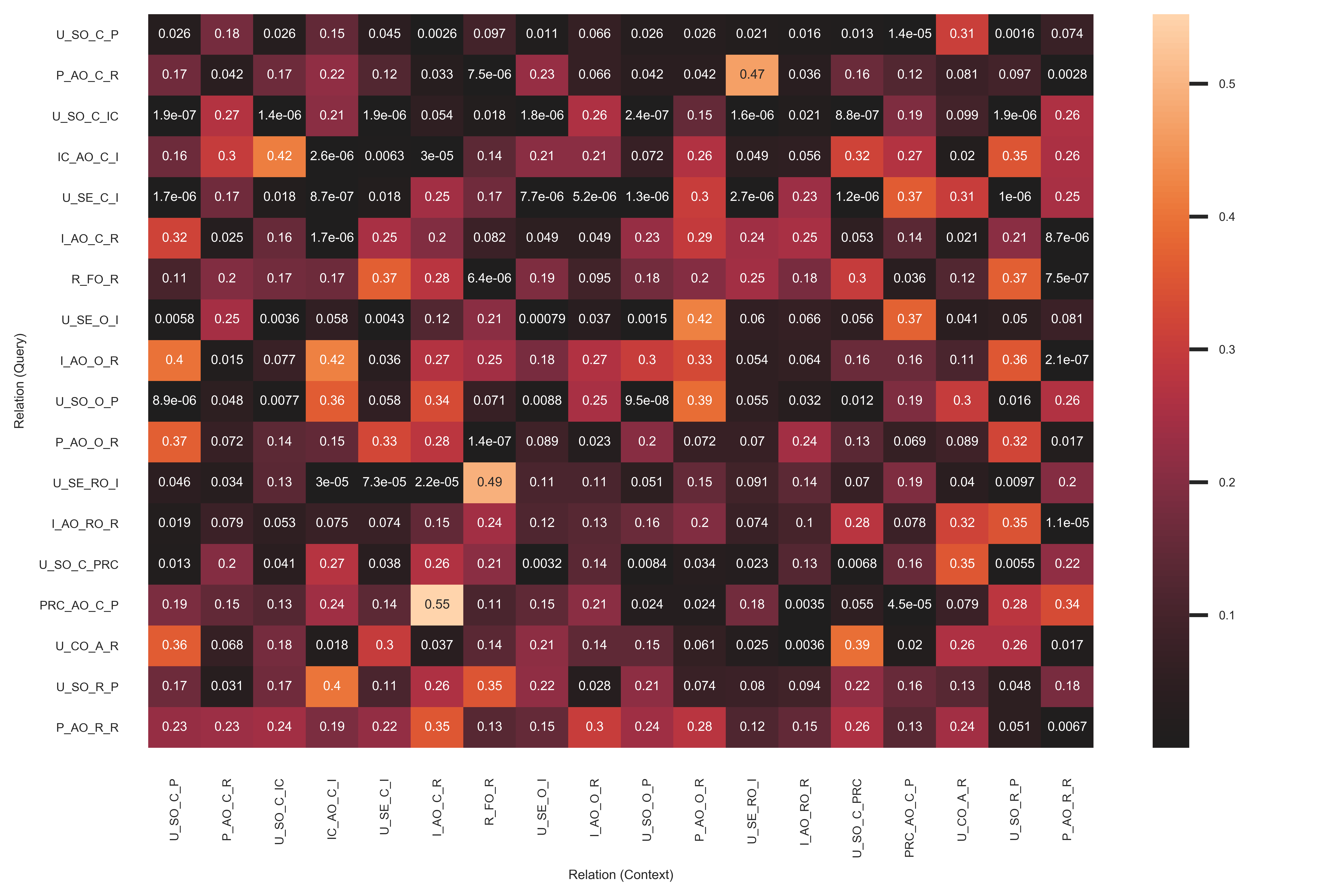}}
\caption{Example heatmap of the absolute importance scores between relations learned as part of the model.}
\label{fig:correlation}
\end{center}
\vskip -0.2in
\end{figure*}

\begin{table*}[ht!]
\caption{Performance comparison on extrapolated Time Prediction.}
\label{tab:experiments_t}
\vskip 0.15in
\begin{center}
\begin{small}
\begin{sc}
\begin{tabular}{lcccccc}
\toprule
\textbf{Dataset}                    & \textbf{Model}        & $\mathbf{HITS@1}$ & $\mathbf{HITS@3}$ & $\mathbf{HITS@10}$    & $\mathbf{MR}$ & $\mathbf{MRR}$    \\
\midrule
\multirow{4}{*}{GITHUB-SE 1Y-REPO}  & RotatE                & 1.77              & 18.27             & 56.05                 & 10.62         & 0.1675            \\
                                    & DE-RotatE             & 3.46              & 7.27              & \textbf{60.73}        & \textbf{9.35} & 0.1724            \\
                                    & RT-DE-RotatE (ours)   & \textbf{6.18}     & \textbf{19.29}    & 55.91                 & 9.40          & \textbf{0.2073}   \\
\cline{2-7}
                                    & Random                & 5.26              & 15.79             & 52.63                 & 9.5           & 0.1867            \\
\bottomrule
\end{tabular}
\end{sc}
\end{small}
\end{center}
\vskip -0.1in
\end{table*}

\subsection{Hyperparameters}
Initially, we tuned our models using the hyperparameter ranges reported in Table \ref{tab:hyperparameters_range} for dropout, $\eta$, $\omega$, and $\alpha$ resulting in total of 72 runs.
Then, following the best hyperparameters achieved on RotatE and DE-RotatE models, we used dropout = $0.4$, $\eta = 0.5$, $\omega = 6.0$, $\alpha = 3 \times 10^{-5}$, $\lambda = 5 \times 10^{-4}$, time-agnostic negative ratio = $256$, time-dependant negative ratio = $32$, batch size = $64$, warm-up steps = $100000$, warm-up $\alpha$ decay rate = $0.1$, steps = $200000$, and validation steps = $10000$ for all experiments.

To make a fair comparison, we chose a base embedding size of 128 for all experiments.
Subsequently, we only report on the combinations of static embedding dimension $d_s$ values and diachronic embedding dimension $d_t$ values presented in Table \ref{tab:hyperparameters_range} where $d_s + d_t = 128$.
We evenly distribute $d_t$ among all diachronic embeddings to prevent giving models a distinct advantage in terms of free-parameters.
As for the relative time embedding dimension $d_r$, we report on all the combinations in Table \ref{tab:hyperparameters_range} with $d_s$ and $d_t$ respecting the stated restriction resulting in total of 17 experiments per dataset.

\subsection{Runtime}
Table \ref{tab:runtime} presents the average runtime of each model for every 100 steps with batch size set to 64. All experiments were carried on servers with 16 CPU cores, 64GB of RAM, and a NVIDIA V100/P100 GPU.

\subsection{Standard Error}
We use standard error to calculate confidence intervals and detect statistically indistinguishable results.

\subsection{Relations Importance Matrix}
Figure \ref{fig:correlation} presents the importance matrix between relations, i.e. $W_P$, learned as part of the RT-X model. From this figure, it is evident that the learned matrix is not symmetric, indicating that the model learns different importance scores conditioned on the query relation.

\subsection{Implementation}
We implemented our model using PyTorch~\cite{paszke2019pytorch}. The source code is publicly available in GitHub\footnote{\url{https://github.com/kahrabian/RT-X}}.

\begin{table*}
\vskip -0.07in
\caption{Extraction rules used to build the KG from raw events.}
\label{tab:extraction}
\vskip 0.1in
\begin{center}
\begin{tiny}
\begin{sc}
\begin{tabularx}{\linewidth}{@{}YYYY@{}}
\toprule
\textbf{Event Type}                                     & \textbf{Head}                                         & \textbf{Relation (Code)}                  & \textbf{Tail}                                         \\ \midrule
Commit Comment                                          & User                                                  & Actor (U\_AO\_CC)                         & Commit Comment                                        \\ \midrule
Fork                                                    & Repository                                            & Fork (R\_FO\_R)                           & Repository                                            \\ \midrule
\multirow{6}{*}{Issue Comment}                          & \multirow{3}{*}{User}                                 & Created (U\_SO\_C\_IC)                    & \multirow{3}{*}{Issue Comment}                        \\ \cline{3-3}
                                                        &                                                       & Edited (U\_SO\_E\_IC)                     &                                                       \\ \cline{3-3}
                                                        &                                                       & Deleted (U\_SO\_D\_IC)                    &                                                       \\ \cline{2-4}
                                                        & \multirow{3}{*}{Issue Comment}                        & Created (IC\_AO\_C\_I)                    & \multirow{3}{*}{Repository}                           \\ \cline{3-3}
                                                        &                                                       & Edited (IC\_AO\_E\_I)                     &                                                       \\ \cline{3-3}
                                                        &                                                       & Deleted (IC\_AO\_D\_I)                    &                                                       \\ \midrule
\multirow{26}{*}{Issues}                                & \multirow{12}{*}{User}                                & Opened (U\_SE\_O\_I)                      & \multirow{12}{*}{Issue}                               \\ \cline{3-3}
                                                        &                                                       & Edited (U\_SE\_E\_I)                      &                                                       \\ \cline{3-3}
                                                        &                                                       & Deleted (U\_SE\_D\_I)                     &                                                       \\ \cline{3-3}
                                                        &                                                       & Pinned (U\_SE\_P\_I)                      &                                                       \\ \cline{3-3}
                                                        &                                                       & Unpinned (U\_SE\_UP\_I)                   &                                                       \\ \cline{3-3}
                                                        &                                                       & Closed (U\_SE\_C\_I)                      &                                                       \\ \cline{3-3}
                                                        &                                                       & Reopened (U\_SE\_RO\_I)                   &                                                       \\ \cline{3-3}
                                                        &                                                       & Assigned (U\_SE\_A\_I)                    &                                                       \\ \cline{3-3}
                                                        &                                                       & Unassigned (U\_SE\_UA\_I)                 &                                                       \\ \cline{3-3}
                                                        &                                                       & Locked (U\_SE\_LO\_I)                     &                                                       \\ \cline{3-3}
                                                        &                                                       & Unlocked (U\_SE\_ULO\_I)                  &                                                       \\ \cline{3-3}
                                                        &                                                       & Transferred (U\_SE\_T\_I)                 &                                                       \\ \cline{2-4}
                                                        & \multirow{2}{*}{User}                                 & Assigned (U\_AO\_A\_I)                    & \multirow{2}{*}{Issue}                                \\ \cline{3-3}
                                                        &                                                       & Unassigned (U\_AO\_UA\_I)                 &                                                       \\ \cline{2-4}
                                                        & \multirow{12}{*}{Issue}                               & Opened (I\_AO\_O\_R)                      & \multirow{12}{*}{Repository}                          \\ \cline{3-3}
                                                        &                                                       & Edited (I\_AO\_E\_R)                      &                                                       \\ \cline{3-3}
                                                        &                                                       & Deleted (I\_AO\_D\_R)                     &                                                       \\ \cline{3-3}
                                                        &                                                       & Pinned (I\_AO\_P\_R)                      &                                                       \\ \cline{3-3}
                                                        &                                                       & Unpinned (I\_AO\_UP\_R)                   &                                                       \\ \cline{3-3}
                                                        &                                                       & Closed (I\_AO\_C\_R)                      &                                                       \\ \cline{3-3}
                                                        &                                                       & Reopened (I\_AO\_RO\_R)                   &                                                       \\ \cline{3-3}
                                                        &                                                       & Assigned (I\_AO\_A\_R)                    &                                                       \\ \cline{3-3}
                                                        &                                                       & Unassigned (I\_AO\_UA\_R)                 &                                                       \\ \cline{3-3}
                                                        &                                                       & Locked (I\_AO\_LO\_R)                     &                                                       \\ \cline{3-3}
                                                        &                                                       & Unlocked (I\_AO\_ULO\_R)                  &                                                       \\ \cline{3-3}
                                                        &                                                       & Transferred (I\_AO\_T\_R)                 &                                                       \\ \midrule
\multirow{3}{*}{Member}                                 & \multirow{3}{*}{User}                                 & Added (U\_CO\_A\_R)                       & \multirow{3}{*}{Repository}                           \\ \cline{3-3}
                                                        &                                                       & Removed (U\_CO\_E\_R)                     &                                                       \\ \cline{3-3}
                                                        &                                                       & Edited (U\_CO\_R\_R)                      &                                                       \\ \midrule
\multirow{6}{*}{Pull Request Review Comment}            & \multirow{3}{*}{User}                                 & Created (U\_SO\_C\_PRC)                   & \multirow{3}{*}{Pull Request Review Comment}          \\ \cline{3-3}
                                                        &                                                       & Edited (U\_SO\_E\_PRC)                    &                                                       \\ \cline{3-3}
                                                        &                                                       & Deleted (U\_SO\_D\_PRC)                   &                                                       \\ \cline{2-4}
                                                        & \multirow{3}{*}{Pull Request Review Comment}          & Created (PRC\_AO\_C\_P)                   & \multirow{3}{*}{Pull Request}                         \\ \cline{3-3}
                                                        &                                                       & Edited (PRC\_AO\_E\_P)                    &                                                       \\ \cline{3-3}
                                                        &                                                       & Deleted (PRC\_AO\_D\_P)                   &                                                       \\ \midrule
\multirow{6}{*}{Pull Request Review}                    & \multirow{3}{*}{User}                                 & Submitted (U\_SO\_S\_PR)                  & \multirow{3}{*}{Pull Request Review}                  \\ \cline{3-3}
                                                        &                                                       & Edited (U\_SO\_E\_PR)                     &                                                       \\ \cline{3-3}
                                                        &                                                       & Dismissed (U\_SO\_D\_PR)                  &                                                       \\ \cline{2-4}
                                                        & \multirow{3}{*}{Pull Request Review}                  & Submitted (PR\_AO\_S\_P)                  & \multirow{3}{*}{Pull Request}                         \\ \cline{3-3}
                                                        &                                                       & Edited (PR\_AO\_E\_P)                     &                                                       \\ \cline{3-3}
                                                        &                                                       & Dismissed (PR\_AO\_D\_P)                  &                                                       \\ \midrule
\multirow{26}{*}{Pull Request}                          & \multirow{12}{*}{User}                                & Assigned (U\_SO\_A\_P)                    & \multirow{12}{*}{Pull Request}                        \\ \cline{3-3}
                                                        &                                                       & Unassigned (U\_SO\_UA\_P)                 &                                                       \\ \cline{3-3}
                                                        &                                                       & Review Requested (U\_SO\_RR\_P)           &                                                       \\ \cline{3-3}
                                                        &                                                       & Review Request Removed (U\_SO\_RRR\_P)    &                                                       \\ \cline{3-3}
                                                        &                                                       & Opened (U\_SO\_O\_P)                      &                                                       \\ \cline{3-3}
                                                        &                                                       & Edited (U\_SO\_E\_P)                      &                                                       \\ \cline{3-3}
                                                        &                                                       & Closed (U\_SO\_C\_P)                      &                                                       \\ \cline{3-3}
                                                        &                                                       & Ready for Review (U\_SO\_RFR\_P)          &                                                       \\ \cline{3-3}
                                                        &                                                       & Locked (U\_SO\_L\_P)                      &                                                       \\ \cline{3-3}
                                                        &                                                       & Unlocked (U\_SO\_UL\_P)                   &                                                       \\ \cline{3-3}
                                                        &                                                       & Reopened (U\_SO\_R\_P)                    &                                                       \\ \cline{3-3}
                                                        &                                                       & Synchronize (U\_SO\_S\_P)                 &                                                       \\ \cline{2-4}
                                                        & \multirow{2}{*}{User}                                 & Assigned (U\_AO\_A\_P)                    & \multirow{2}{*}{Pull Request}                         \\ \cline{3-3}
                                                        &                                                       & Unassigned (U\_AO\_U\_P)                  &                                                       \\ \cline{2-4}
                                                        & \multirow{2}{*}{User}                                 & Review Requested (U\_RRO\_A\_P)           & \multirow{2}{*}{Pull Request}                         \\ \cline{3-3}
                                                        &                                                       & Review Request Removed (U\_RRO\_R\_P)     &                                                       \\ \cline{2-4}
                                                        & \multirow{12}{*}{Pull Request}                        & Assigned (P\_AO\_A\_R)                    & \multirow{12}{*}{Repository}                          \\ \cline{3-3}
                                                        &                                                       & Unassigned (P\_AO\_UA\_R)                 &                                                       \\ \cline{3-3}
                                                        &                                                       & Review Requested (P\_AO\_RR\_R)           &                                                       \\ \cline{3-3}
                                                        &                                                       & Review Request Removed (P\_AO\_RRR\_R)    &                                                       \\ \cline{3-3}
                                                        &                                                       & Opened (P\_AO\_O\_R)                      &                                                       \\ \cline{3-3}
                                                        &                                                       & Edited (P\_AO\_E\_R)                      &                                                       \\ \cline{3-3}
                                                        &                                                       & Closed (P\_AO\_C\_R)                      &                                                       \\ \cline{3-3}
                                                        &                                                       & Ready for Review (P\_AO\_RFR\_R)          &                                                       \\ \cline{3-3}
                                                        &                                                       & Locked (P\_AO\_L\_R)                      &                                                       \\ \cline{3-3}
                                                        &                                                       & Unlocked (P\_AO\_UL\_R)                   &                                                       \\ \cline{3-3}
                                                        &                                                       & Reopened (P\_AO\_R\_R)                    &                                                       \\ \cline{3-3}
                                                        &                                                       & Synchronize (P\_AO\_S\_R)                 &                                                       \\ \midrule
Push                                                    & User                                                  & Sender (U\_SO\_C)                         & Repository                                            \\ \midrule
\multirow{2}{*}{Star}                                   & \multirow{2}{*}{User}                                 & Created (U\_HS\_A\_R)                     & \multirow{2}{*}{Repository}                           \\ \cline{3-3}
                                                        &                                                       & Deleted (U\_HS\_R\_R)                     &                                                       \\
\bottomrule
\end{tabularx}
\end{sc}
\end{tiny}
\end{center}
\vskip -0.18in
\end{table*}

\end{document}